World Conference of Transport Research, Toulouse 2026 (WCTR 2026 Toulouse)

# PAPN: Proximity Attention Encoder and Pointer Network Decoder for Parcel Pickup Route Prediction

Hansi Denis*[a], Ali Anwar[a], Ngoc-Quang Luong[b], Siegfried Mercelis[a]

[a]*IDLab, Faculty of Applied Engineering, University of Antwerp-imec, Sint-Pietersvliet 7, 2000 Antwerp, Belgium*
[b]*Formerly with University of Antwerp – imec Antwerp, Belgium*

**Abstract**

Optimization of the last-mile delivery and first-mile pickup of parcels is integral to the logistics optimization pipeline as it entails both cost and resource efficiency and a heightened service quality. Such optimization requires accurate route and time prediction systems to adapt to different scenarios in advance. This work tackles the first building block, namely route prediction. The novel Proximity Attention (PA) mechanism is coupled to a Pointer Network (PN) decoder to leverage the underlying connections between the different visitable pickup positions at each timestep of the parcel pickup process. This local attention is coupled with global context computing via a multi-head attention transformer encoder. Both attentions are then mixed for complete and comprehensive modeling of the problems. PA is also used in the decoding process to skew predictions towards the locations with the highest visit likeliness, thus using inter-connectivity of nodes for next-location prediction. This method is trained, validated and tested on a large industry-level dataset of real-world, last-mile delivery and first-mile pickup named LaDE (2024). This approach outperforms all state-of-the-art supervised methods in terms of most metrics used for benchmarking on this dataset while still being competitive with the best-performing reinforcement learning framework named DRL4Route (2023).

## 1. Introduction

Route prediction at its core consists in predicting the path an entity will take from a departure location to a destination location. The route to be predicted can have a predefined set of locations to visit. Furthermore, if working in an environment where time is kept, time constraints can be set on when certain locations are visitable. In that case, the problem at hand is a time-window constrained route prediction, the problem dealt with in this study. In the case of last-mile delivery and first-mile pick-up, this entity can be a delivery or pick-up person operating various kinds of vehicles (trucks, scooters, bikes, or others) or autonomous devices such as rolling or flying drones. For this study, the travelling entities are human personnel riding scooters as the work is conducted and evaluated on the LaDE(2024) dataset. It is noteworthy that this work predicts that route which is most likely to be selected by the carrier, and not the theoretically optimal one. This is inspired by the fact that, contrary to programmed drones, the route used to reach the destination location is heavily linked to human judgement as traffic, weather conditions or road network knowledge can influence the order in which the locations are visited. In addition, such predictions are instrumental to various extent: from evaluating the effect of certain carriers' activity on the global urban traffic, heightening customer satisfaction, assisting order dispatching, to serving as tutor for training new delivery/pickup personnel as highlighted in Wen *et al.*(2024) survey on this problem .

* Corresponding author. Tel.: +33-6-43-06-38-76.
*E-mail address:* hansi.denis@uantwerpen.be

In this paper, a novel approach to learning route prediction in the context of first-mile parcel pickup is proposed. This approach builds upon the axis studied by Wen *et al.*(2022) of implementing a graph-based encoder-decoder architecture leveraging the multi-head attention developed by Vaswani *et al.*(2023) and Vinyals *et al.*(2015)'s pointer network. The main innovative component is the double use of proximity attention for both location attention update and skewing of the node selection process. Proximity attention consists of computing attention over the reachability mask of the visitable locations. Leveraging this aspect of the data permits more reliable route prediction for the full length of the route. This ensures a more stable route prediction that captures both spatial proximity and global state of the further to-be-visited nodes. A balance is then struck between the immediate context provided by the proximity attention and the global context given by the transformer encoder. A good example of the benefits of this approach can be given where we consider a pickup 1 kilometer away (due to being picked up in 2 hours) and another 10 kilometers away (due to being picked up in 15 minutes). In that case, proximity attention groups the former with adjacent nodes, while global attention overrides spatial bias to prioritize the latter. Training the model with both considerations leads to better decision making in the route building process. Results show that this novel approach boasts a comparatively higher performance with respect to the existing supervised models in the literature. The summary of the contributions in this paper is the following:

(1) Building of a supervised encoder-decoder architecture for stable route prediction.

(2) Implementation of proximity attention, leveraging availability of nodes at each prediction step to produce attention weights for node attention update and node selection in the route-building process. This gives way to the first implementation of hybrid attention mechanism for first-mile pickup logistics. Leveraging top-k masking and global attention ensures balanced consideration of local route cohesion and global state responsiveness.

(3) Comparative and qualitative studies with respect to the influence of proximity attention at both steps as well as prediction quality.

The rest of this paper is structured as follows: Section 2 depicts the existing methods in the literature employed to tackle similar problems. Section 3 consists in a thorough explanation of the methodology used, by clearly defining each component of the model both structurally and mathematically. Section 4 serves as an exposition of experimental settings including the dataset used, the resources employed as well as the baseline and metrics used for evaluation. The results and comparative and qualitative studies for better insights into this novel approach's impact make up Section 5. Finally, this study is concluded in Section 6 with a summary of the content along with perspectives for future work.

## 2. Literature review

Last and first-mile routing problems are extensively researched in the literature, be it from optimization, emission estimation, route prediction, arrival time prediction, or traffic overseeing perspectives. All those aspects of routing problems are intricately linked and offer a high number of research axes. In the present case, focus is on route prediction in the context of first mile pick up of parcels with time windows for location availability meaning the timeframe where the parcel should be picked up. The subject of this paper being leveraging supervised learning to tackle this problem, it is interesting to look at vehicle routing problems in general as one can assume that given an optimal data set the method described in this paper could be adapted for optimal routing problems. In the current case the learned routing policy is based on historical data.

In Liu *et al*.(2023)'s work, a systematic literature review of vehicle routing problems with time windows is proposed. The main takeaways are that at the moment of writing, most methods to tackle such problems are either approximate methods such as meta-heuristics or hybrid algorithms combining multiple approaches. Furthermore, the use of deep learning algorithms for such problems is shown to be widely underutilized mainly because of lack of publicly available large-scale datasets in that research context.

As far as first-mile pickup is concerned, some studies were conducted to evaluate the disruption of new incoming requests on the existing routing plan as in Giménez-Palacios *et al*.(2022). This study, like others, relied on synthetic data that did not accurately represent real-world situations.

For last mile delivery and first mile pickup more specifically, recent studies have made use of deep learning to predict and optimize routing. Such as Yan *et al*.(2024), where the use of attention from Vaswani *et al.*(2023) in the encoding process as well as resource aware allocation of couriers to different tasks is very promising although the use of reinforcement learning can raise the training resources questions. However, the data they use to train their model is not made publicly available and as such does not permit easy performance evaluation and comparison with other existing methods.

To circumvent this lack of comparability in the space of first mile pick up routing prediction, Wu *et al.* released LaDE (2024) publicly. LaDE is a large-scale real-world dataset provided by the Chinese logistics company Cainiao. "LaDe has three unique characteristics: (1) Large-scale. It involves 10,677k packages of 21k couriers over 6 months of real-world operation. (2) Comprehensive information. It offers original package information, task-event information, as well as detailed couriers' trajectories and road networks. (3) Diversity. The dataset includes data from various scenarios, including package pick-up and delivery, and from multiple cities, each with its unique spatial-temporal patterns due to their distinct characteristics such as populations".

The volume of data and the quality of the underlying information contained is well suited for the field of first mile pickup route prediction as it enables the implementation of deep learning methods on a publicly available dataset and thus benchmarking and comparison of different methods.

The authors of this work also released the dataset generation code publicly which yields a structured (sequence or graph) dataset which is particularly fit for route prediction problems. This ensures thorough comparability of results as the datasets used by the different models stem from the exact same root. Since the release of this data set, multiple methods have been implemented, tested and evaluated on it, with the most prominent ones being cited thereafter.

Soon after the release of the dataset Graph2Route (2022) was proposed by Wen *et al.*, which differed from then existing sequence-based methods like Wen *et al.*'s DeepRoute (2022) by its dynamic spatial-temporal graph-based structure. It set the state-of-the-art at the time of release but lacked explicit leveraging of the mask information in the encoding process.

Not long after, Cai *et al.* released M2G4RTP: A Multi-Level and Multi-Task Graph Model for Instant-Logistics Route and Time Joint Prediction (2023). This method achieves joint route and time prediction by using a then novel GAT-e module and achieves state-of-the-art performance in both target tasks. From a route prediction perspective, results show high performance in the first few nodes of the route but relatively low performance maintenance over the rest of the prediction sequence.

Based on this literature review, it was observed that there is a need for more accurate route prediction methods and that leveraging node availability in the encoding process could be beneficial to use local context in both encoding and decoding steps to obtain the desired results. From this observation comes the novelty brought in this work of embedding the node masking information and effectively incorporating it in the encoding process of each instance. The encoder-decoder architecture is kept like Graph2Route (2022) and M2G4RTP (2023) but this work differs from them by having two subsequent encoding layers operating on two different levels of context modelling. This balances early-step accuracy and long-sequence robustness. Moreover, in existing methods the masking information is only leveraged in the decoding process to reduce the potential nodes to be selected whereas two-level exploitation of the availability of nodes is used here, namely in encoding (proximity attention with Top-k) and in decoding (probability-level mask addition), while training purely supervised .

## 3. Methodology

This section describes the methodology employed via extensive structural and mathematical architecture description of the PAPN model. Fig.1 introduces the general workflow of the model; it is designed using an encoder (Section 3.2-4)-decoder (Section 3.5) to produce route predictions. The encoder receives the node and edge features along with the reachability mask and embeds them in a higher dimensional space. Those embeddings then pass through two encoding layers (proximity attention and transformer encoder from DeepRoute (2022)) which respective outputs are then merged. The obtained representation is fed to the decoder which then produces next-node prediction for each timestep leveraging a combination of Pointer Network, LSTM and the reachability mask attention computed in the encoder.

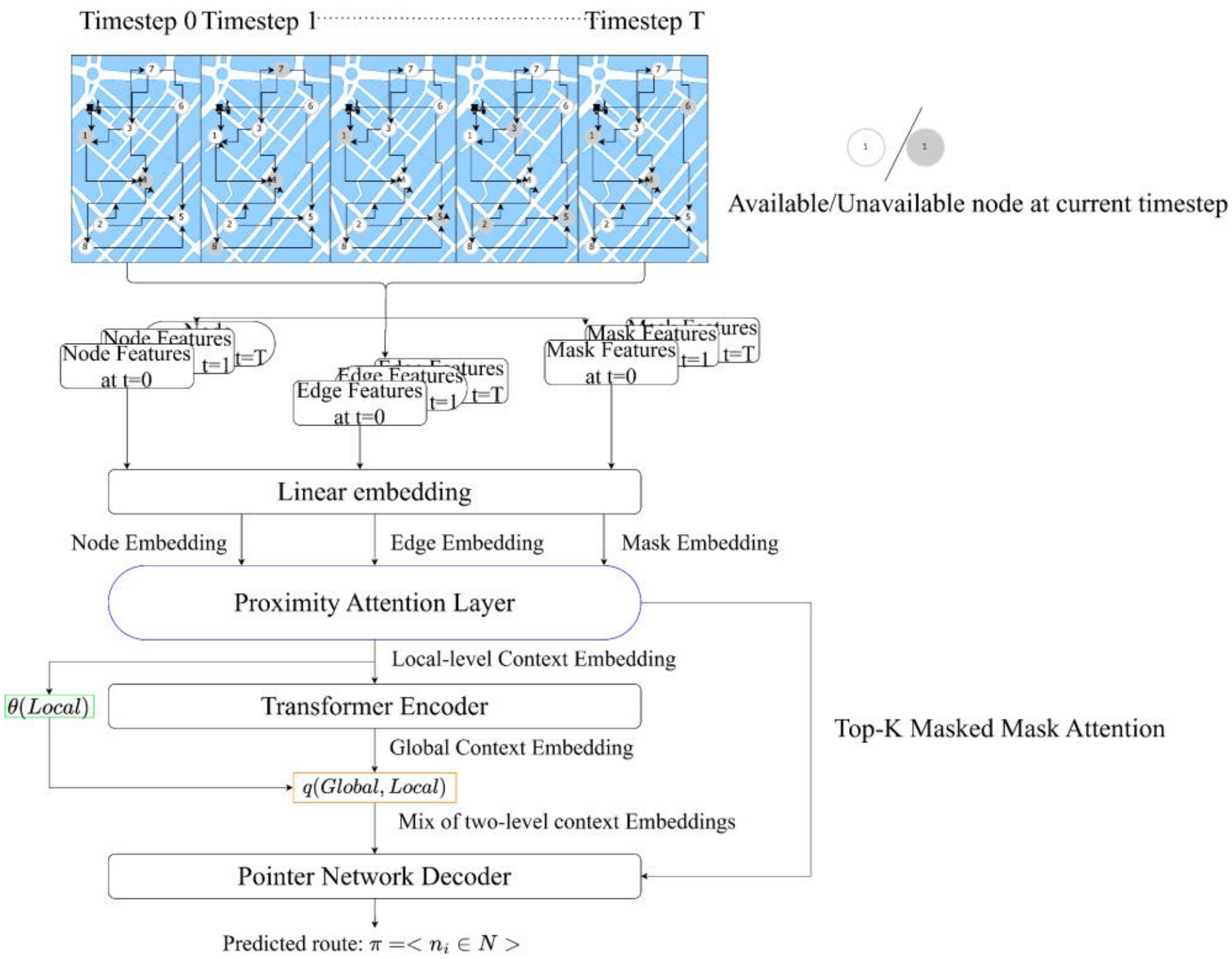

Fig. 1. PAPN Workflow

### 3.1. Input Data

The node features used in this work are the following: the time at which the task was accepted, longitude and latitude of the node, distance of the node to the starting point, relative distance with respect to the courier's average travel distance per day, time difference between acceptance time of the task and promised time of parcel pickup and features of the area of interest in which the node is situated. Including its ID, type, and position. Feature representation is kept the same as in Graph2Route (2022) with the exception of the area of interest information which is concatenated to the node data to add neighborhood context.

### 3.2. Linear Embedding Layer

As a first step after receiving the input data, a linear embedding layer is applied to all input features. It permits a constant consistent scaling of the data to be used in the following modules. For further merging of the data, all three components, meaning nodes, edges, and mask, are embedded into the same dimensional hidden space. The linear layer applied conforms to the following expression:

$$y = xA + b \tag{1}$$

Where x is the input feature vector, A is the weight matrix that reshapes the embedding to the hidden size and b is the bias vector.

### 3.3. Proximity Attention

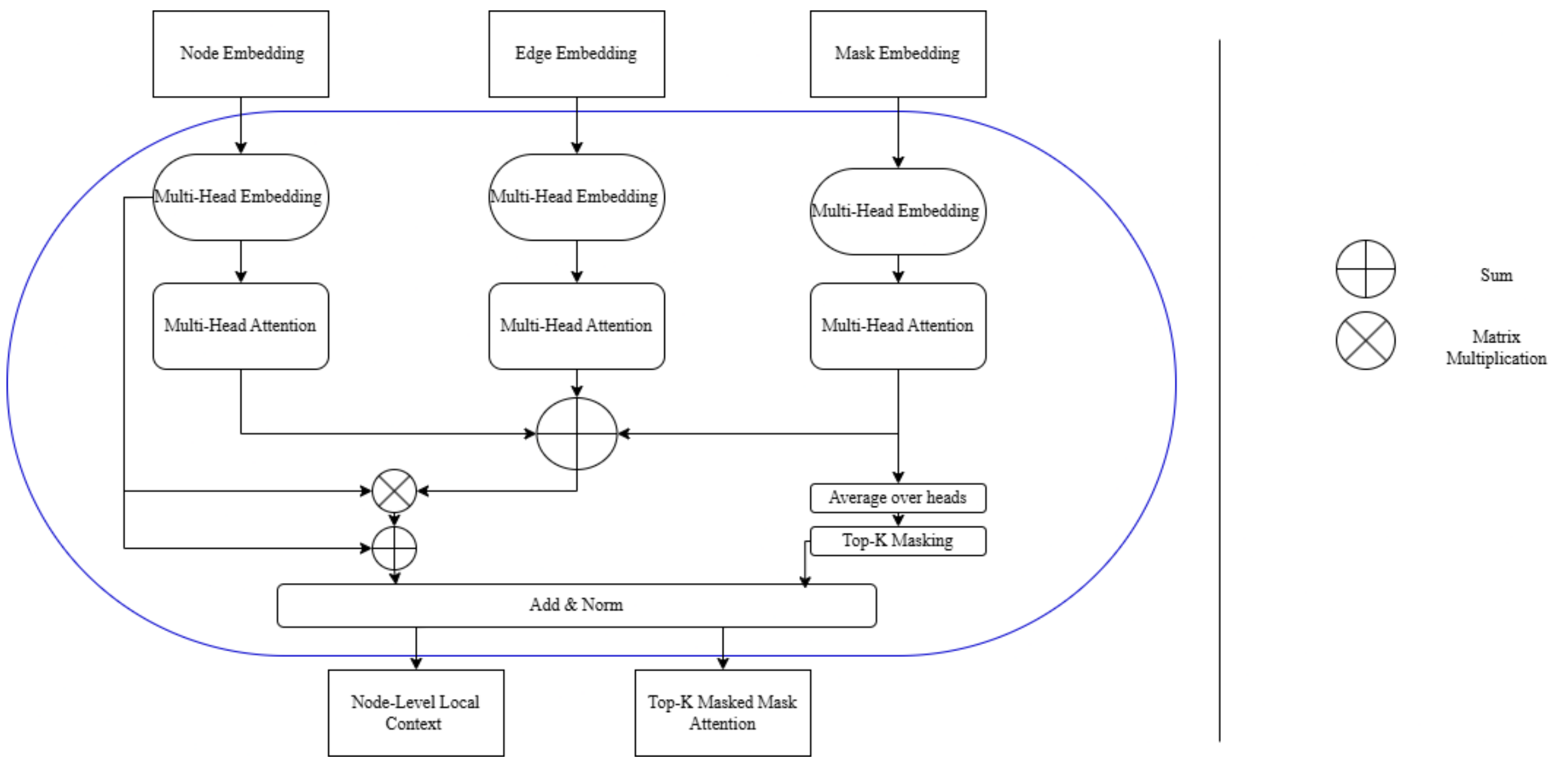

Fig. 2. Proximity Attention Layer

The Proximity Attention Layer is illustrated in Fig.2, which constitutes the major novelty brought by this paper, enhances feature representations by jointly modelling their interactions through multi-head attention. This enables the capture of local dependencies between the nodes by leveraging edge and mask information without sacrificing special constraints. The node features are updated by merging them with the attention computed over the edge and mask, thus aggregating local information on a node level. First, the input features are flattened to accommodate head-wise representation. The obtained $\tilde{X}$ , $\tilde{E}$ and $\tilde{M}$ are then projected to multi-head space by using learnable matrices $W$ , $W_e$ and $W_m$:

Projection:

$$X' = \tilde{X} \cdot W \in R^{N\times\left(H\cdot D_x^{'}\right)}, E' = \tilde{E} \cdot W_e \in R^{N\times N\times\left(H\cdot D_e^{'}\right)},\ M' = \tilde{M} \cdot W_m \in R^{N\times\left(H\cdot D_m^{'}\right)} \tag{2}$$

Proximity attention layer computes multi-head attention scores:

$$\alpha_{ij}^h = SoftMax\left(\text{LeakyReLU}\left(W_{h_i} X_i^h + W_{h_j} X_j^h + W_e E_{ij}^h + W_{m_i} M_i^h + W_{m_j} M_j^h\right)\right) \tag{3}$$

using node, edge, and mask projections. The attention scores for each head are derived from $X_i^h \in R^{\left(n\times D_x^{'}\right)}$, $E_{ij}{}^h \in R^{\left(n\times n\times D_e^{'}\right)}$, and $M_i^h \in R^{\left(n\times D_m^{'}\right)}$. To help measure the influence of the mask attention in the downstream selection of nodes for route construction, a k is chosen arbitrarily to perform top-k filtering on the mask attention to filter the info only of the nodes with the most significant attention scores:

$$\left(\text{TopKMask} = \text{multi_hot}\left(\text{top}_k\left(m_{\text{att}i}\right)\right)\right) \tag{4}$$

Where the mask attention $m_{att}$ refers to the multi-head attention computed based on the mask embedding. The resulting filtered mask attention is the computed attention to which is applied the top-k operator resulting in a vector of the same length as the attention one but with zeros in places filtered out.

$$\left(\boldsymbol{M} = \boldsymbol{M} \circ TopKMask\right) \tag{5}$$

Finally, node features are updated by adding the SoftMax-normalized attention to the input node features and the mask attention is updated by adding the SoftMax-normalized mask attention to the input mask features.

### 3.4. Transformer Encoder

After passing through the proximity attention layer, the node features with added context from the edge and mask attention is fed to a transformer encoder leveraging multi-head attention to obtain a global context representation of the instance and proceed to the decoding process. The multi-head attention leveraged in this encoder is formulated in Vaswani *et al*. (2023) and the transformer encoder used is akin to the one in DeepRoute (2022).

### 3.5. Attention Mixing

After producing global-level embedding, attention mixing is performed to combine it with local-level embedding and get context that fully grasp both levels of representations as an input for the last module of the model. As local context and global context have different dimensions, aggregation is done over the proximity attention layer's output to match the global context's dimension. Multiple aggregation methods, namely sum, mean, minimum, and maximum, are experimented on. The general notation for the dimension reduction via aggregation is:

$$L_{\mathrm{ag}} = \theta\left(L\right) \tag{6}$$

With $L$ being the local embedding, $L_{\mathrm{ag}}$ being its aggregated version and being the aggregation function used. More specifically, aggregation is done on the second dimension of this embedding which represents the node level information dimension noted $d_{nodes}$. The following formulations for the different aggregation functions are obtained with $i$ indexing timesteps, $j$ indexing nodes, and $k$ indexing the dimensions of the local feature embedding. The addition operator sums the proximity attention over the nodes dimension $j$ :

$$\theta_{\mathrm{sum}}\left(L_{i,k}\right) = \sum_{j=1}^{d_{nodes}} L_{i,j,k} \tag{7}$$

The average reduction function averages the proximity attention for each timestep over the nodes dimension:

$$\theta_{\mathrm{mean}}\left(L_{i,k}\right) = \frac{1}{d_{local}} \sum_{j=1}^{d_{nodes}} L_{i,j,k} \tag{8}$$

Maximum and minimum reduction respectively keep the maximum or minimum value for each timestep among the different nodes:

$$\theta_{\text{max or min}}\left(L_{i,k}\right) = \text{max or min}_{j}\left(L_{i,j,k}\right) \tag{9}$$

After dimension reduction, both attentions are mixed using sum or random selection of values from each embedding. Both aggregation and mixing steps are followed by applying normalization to the results. Let $q$ combine the global context $G$ and the aggregated local context $\mathrm{L}_{ag}$ into a mixed context $C \in \mathbb{R}^{T \times d_{nodes}}$ . Sum mixing is defined as:

$$C = q_{sum}\left(G, \mathrm{L}_{ag}\right) = \mathrm{Norm}\left(G + \mathrm{L}_{ag}\right). \tag{10}$$

Random selection mixing involves defining a Bernoulli mask $M \in \{0,1\}^{T \times d_{nodes}}$ sampled elementwise with a set replacement probability $p$ and let $\odot$ denote elementwise product. Then:

$$M_{i,k} \sim \text{Bernoulli}(p) \quad (11)$$

$$C = q_{rand}(G, \text{L}_{ag}) = \text{Norm}(M \odot G + (1-M) \odot \text{L}_{ag}) \quad (12)$$

*3.6. Pointer Network Decoder*

The final module of the model is the Pointer Network Decoder; its function is to autoregressively predict the node sequence by leveraging attention over encoder-received outputs and proximity attention layer originating mask attention. The version used for this work is an adaptation of the one multi-level decoder used in M2G4RTP(2023) to cater to solely route prediction. The Pointer Network is instrumental in handling output sequences of varying lengths as not all instances contain the same number of locations to be visited. For each timestep, a combination of LSTM as formulated in Staudemeyer and Morris (2019), attention-based glimpses, and pointer attention is leveraged to calculate logits for each node while respecting inherent mask constraints, meaning the unavailable nodes at each timestep. The main adaptation brought in this work is the use of mask attention (top-k masked or not) computed in the proximity attention layer in addition to the probability to skew the selection process towards the most relevant nodes local-context wise:

$$P(\pi_t = i) = \text{Softmax}(u_{ti} + M_{att}) \quad (13)$$

Where $(M_{att})$ contains the mask attention and masks unavailable nodes while $u_{ti}$ constitutes the logit score for node $i$ at timestep $t$. As the output of the model, the routing sequence predicted by the model is obtained as a reordering of the nodes-to-be-visited list in the input. We obtain a routing sequence :

$$\pi = < n_i \in N > \quad (14)$$

## 4. Experiments and outcomes

This section describes the experimental environment used to evaluate and benchmark PAPN against existing methods.

*4.1. Dataset*

The dataset used for this study is LaDE (2024) dataset. This dataset is an industry level dataset of delivery and pickup scenarios for parcels in the major Chinese cities. The training and evaluation of this model was done in the city of Hangzhou, which is the one with the most available data. The dataset used is a graph dataset with node and edge features along with a reachability mask for each timestep. The experiments are conducted on a real-world package pick-up dataset. The time-range covered is 6 months in 2021. This represents in total 10,667k packages and 619k trajectories generated by 21k couriers, covering 5 cities over a total span of 6 months. The maximal package number of a courier one trip in the pick-up scenario reaches 95 as reported in LaDE (2024). For a realistic comparison of the proposed method compared to the existing ones, the only filtering performed on the instances is that only the ones with less than 25 tasks to be performed are kept, resulting in 222699 instances. To add area-of-interest (aoi) information to the node raw features, aoi node data is concatenated with the existing node data.

*4.2. Settings*

The experiments are run on JupyterHub using two CPUs with 20 GB of memory each and a 1080TI GPU. The hidden dimension parameter is set at 128 for all methods. Batch size is set to 64. For PAPN, heads in Transformer are set to 8. The Transformer Encoder has two layers and there is one Proximity Attention layer.

*4.3. Baselines*

The proposed approach is put against a set of existing methods for comparison; these methods are evaluated on the same dataset as PAPN and follow the same exact experimentation protocol as described in section 4.1 of LaDE (2024), ensuring a fair performance comparison. The reported metrics for the baselines reported hereafter are taken directly from LaDE (2024). The methods chosen as baselines are the following:

- Distance-Greedy, a greedy method of ordering the nodes based on the distance separating them.
- Perron and Furnon's OR-Tools (2025), a heuristic method for shortest-path finding.
- OSquare, which leverages Ke *et al.'s* XGBoost(2019) to recurrently predict route.
- DeepRoute (2022), which leverages a Transformer-based encoder-decoder architecture to rank the packages to be picked up.
- Graph2Route (2022), models the locations to be visited as a graph and utilizes a graph convolutional network-based encoder and attention decoder to predict the route travelled.
- M2G4RTP (2023) which leverages multi-level task representation and aoi information to perform joint route and time prediction.
- DRL4Route (2023), a deep reinforcement learning framework for route prediction to be used as anvironment free add-on on top of an existing deep learning model. Not a stand-alone approach. The results reported for this method are those of the framework applied to a DeepRoute (2022)-like architecture.

### *4.4. Metrics*

For evaluation purposes, comparison between the predicted route and the real route is made by computing LSD (Location Square Deviation), HR@k, KRC (Kendall Rank Correlation) and ED (Edit Distance). Those metrics were provided along with their implementation in the code repository for the LaDE (2024) dataset, ensuring consistency in the way models are evaluated throughout the benchmarking process.

### *4.5. Results*

Table 1. Benchmarking results.

| Method | Chongqing | | | | Shanghai | | | | Yantai | | | |
|---|---|---|---|---|---|---|---|---|---|---|---|---|
| | ***HR@3*** | ***KRC*** | ***LSD*** | ***ED*** | ***HR@3*** | ***KRC*** | ***LSD*** | ***ED*** | ***HR@3*** | ***KRC*** | ***LSD*** | ***ED*** |
| **Time-Greedy** | 63.86 ±0.00 | 44.16 ±0.00 | 3.91 ±0.00 | 1.74 ±0.00 | 59.81 ±0.00 | 39.93 ±0.00 | 5.20 ±0.00 | 2.24 ±0.00 | 61.23 ±0.00 | 39.64 0.00 | 4.62 ±0.00 | 1.85 0.00 |
| **Distance-Greedy** | 62.99 ±0.00 | 41.48 ±0.00 | 4.22 ±0.00 | 1.60 ±0.00 | 61.07 ±0.00 | 42.84 ±0.00 | 5.35 ±0.00 | 1.94 ±0.00 | 62.34 ±0.00 | 40.82 ±0.00 | 4.49 ±0.00 | 1.64 ±0.00 |
| **Or-Tools (2025)** | 64.19 ±0.00 | 43.09 ±0.00 | 3.67 ±0.00 | 1.55 ±0.00 | 62.50 ±0.00 | 44.81 ±0.00 | 4.69 ±0.00 | 1.88 ±0.00 | 63.27 ±0.00 | 42.31 ±0.00 | 3.94 ±0.00 | 1.59 ±0.00 |
| **LightGBM (2017)** | 71.55 ±0.00 | 54.53 ±0.00 | 2.63 ±0.00 | 1.54 ±0.00 | 70.63 ±0.00 | 54.48 ±0.00 | 3.27 ±0.00 | 1.92 ±0.00 | 70.41 ±0.00 | 52.90 ±0.00 | 2.87 ±0.00 | 1.59 ±0.00 |
| **Gao *et al.(*2021)** | 69.98 ±0.32 | 52.07 ±0.38 | 3.36 ±0.04 | 1.51 ±0.01 | 69.05 ±1.23 | 52.72 ±1.72 | 4.08 ±0.25 | 1.86 ±0.03 | 69.08 ±0.61 | 50.62 ±1.20 | 3.60 ±0.15 | 1.57 ±0.02 |
| **DeepRoute (2022)** | 72.09 ±0.39 | 55.72 ±0.40 | 2.66 ±0.06 | 1.51 ±0.01 | 71.66 ±0.10 | 56.20 10.23 | 3.26 ±0.07 | 1.86 ±0.01 | 71.44 ±0.28 | 54.74 ±0.49 | 2.80 ±0.02 | 1.53 ±0.02 |
| **Graph2Route (2022)** | 72.31 ±0.20 | 56.08 ±0.14 | 2.53 ±0.01 | 1.50 ±0.01 | 71.69 ±0.10 | 56.53 ±0.10 | 3.12 ±0.01 | 1.86 ±0.00 | 71.52 ±0.14 | 55.02 ±0.10 | 2.71 ±0.01 | 1.54 ±0.00 |
| **M2G4RTP (2023)** | 72.44 ±0.11 | 56.15 ±0.04 | 2.55 ±0.03 | 1.50 ±0.00 | 71.73 ±0.06 | 56.34 ±0.08 | 3.16 ±0.02 | 1.86 ±0.01 | 71.73 ±0.01 | 55.02 ±0.17 | 2.78 ±0.00 | 1.53 ±0.01 |
| **PAPN** | 72.79 ±0.06 | 56.20 ±0.03 | 2.61 ±0.05 | **1.47 ±0.01** | 71.95 ±0.15 | 56.80 ±0.10 | 3.14 ±0.02 | 1.84 ±0.02 | 71.32 ±0.11 | 54.6 ±0.20 | 2.80 ±0.15 | 1.52 ±0.01 |
| **PAPN-Mixed** | 72.85 ±0.10 | 56.50 ±0.02 | 2.56 ±0.03 | **1.47 ±0.01** | 72.12 ±0.10 | 57.00 ±0.10 | 3.12 ±0.02 | **1.83 ±0.01** | 71.81 ±0.13 | 55.40 ±0.11 | 2.71 ±0.02 | **1.51 ±0.01** |
| **DRL4Route (2023)** | **73.12 ±0.06** | **57.23 ±0.12** | **2.43 ±0.01** | 1.48 ±0.01 | **72.18 ±0.15** | **57.20 ±0.18** | **3.06 ±0.02** | 1.84 ±0.01 | **72.07 ±0.06** | **55.94 ±0.10** | **2.62 ±0.00** | **1.51 ±0.00** |

The results of the baselines from literature and our newly developed PAPN method can be found in Table I for comparison purposes. Results in bold indicate highest performance and results underlined indicate second-best performance. Regarding performance with this set of hyperparameters, a claim can be made that PAPN is the best performing supervised method and standalone method altogether (DRL4Route (2023) is a framework that leverages reinforcement learning in addition to an existing deep learning model). Furthermore, PAPN shows time efficiency compared to DRL4Route, indeed peak performance is attained on average in 8 training epochs for PAPN whereas DRL4Route(2023) report attaining max reward in the 15 to 20 epochs range in their work. This is all considering that on the hardware used for experiments (NVIDIA GeForce GTX 1080 Ti and 2 CPUs with 20GB of memory each), PAPN training epochs last approximately 5 minutes whereas DRL4Route (2023)'s last around 11 minutes.

Indeed, PAPN with mixed embeddings (results shown refer to the best combination of aggregation and mixing methods in term of KRC) outperforms all supervised baselines in most of the evaluation metrics for the Chongqing and Shanghai datasets and outperforms all methods DRL4Route included in terms of ED for these two same datasets. The lower comparative performance on the Yantai dataset outside of ED can be attributed to the fact that Yantai is a smaller city with less orders and a simpler road network thus limiting the amount of information the model can leverage when computing local dependencies between locations. As for PAPN without mixing embeddings, the observed performance is less than with the mixed embeddings but is still competitive with respect to supervised state-of-the-art methods. The results support the use of an embedding mixing step in the model.

### *4.6. Ablation and Leaning rate studies*

The literature shows that a lot of methods developed in the context in which this work is conducted opt for an encoder-decoder architecture. To differentiate this work from the others, proximity attention was introduced to perform two levels of embeddings for a more precise prediction. This added component deepened the model and now poses the question of purpose and usefulness of this novelty. Furthermore, the deeper nature of the model also gives way to interrogation regarding the appropriate learning rate to be used for this model to avoid overfitting. To answer those questions, the following part constitutes an ablation study followed by a study of the impact of the learning rate on performance. To conduct the ablation study, first the essential components of the architecture are determined. It was determined that the components that could potentially be bypassed are both the proximity attention layer and the transformer encoder. The choice was made to study the bypassing of the transformer encoder that produces the global level encoding to measure the ability of the local embeddings to perform by themselves as well as whether producing both levels of embeddings and mixing them is beneficial. Once again, for a fair comparison, the hyperparameters used are the same as in section 4.2, and the aggregated version of the local embedding is directly input into the decoder. Comparison of the results obtained by using only proximity attention-based embeddings (denoted OPAPN) with the ones from vanilla PAPN with mixed embeddings are in Table 2. One can observe that vanilla PAPN outperforms the version stripped of global embedding consideration in all metrics thus supporting the idea that combining two levels of information is beneficial performance wise.

Table 2. Ablation and learning rate studies result.

| Method | Chongqing | | | | Shanghai | | | | Yantai | | | |
|---|---|---|---|---|---|---|---|---|---|---|---|---|
| | ***HR@3*** | ***KRC*** | ***LSD*** | ***ED*** | ***HR@3*** | ***KRC*** | ***LSD*** | ***ED*** | ***HR@3*** | ***KRC*** | ***LSD*** | ***ED*** |
| **OPAPN** | 72.71 ±0.03 | 56.34 ±0.07 | 2.58 ±0.05 | 1.48 ±0.01 | 72.03 ±0.11 | 56.90 ±0.10 | 3.15 ±0.03 | 1.84 ±0.02 | 71.6 ±0.10 | 54.80 ±0.20 | 2.82 ±0.09 | 1.53 ±0.01 |
| **PAPN -Mixed** | **72.85 ±0.10** | **56.50 ±0.02** | **2.56 ±0.03** | **1.47 ±0.01** | **72.12 ±0.10** | **57.00 ±0.10** | **3.12 ±0.02** | **1.83 ±0.01** | **71.71 ±0.13** | **54.94 ±0.11** | **2.77 ±0.02** | **1.52 ±0.01** |
| **PAPN-Mixed FT** | 72.90 ±0.10 | 56.70 ±0.04 | 2.52 ±0.02 | **1.47 ±0.01** | **72.21 ±0.10** | 57.10 ±0.10 | 3.10 ±0.01 | **1.83 ±0.01** | 71.80 ±0.08 | 55.41 ±0.11 | 2.71 ±0.02 | 1.51 ±0.01 |
| **DRL4Route (2023)** | **73.12 ±0.06** | **57.23 ±0.12** | **2.43 ±0.01** | 1.48 ±0.01 | 72.18 ±0.15 | **57.20 ±0.18** | **3.06 ±0.02** | 1.84 ±0.01 | **72.07 ±0.06** | **55.94 ±0.10** | **2.62 ±0.00** | **1.51 ±0.00** |

As for the impact of an arbitrary learning rate for benchmarking, a study is conducted to search for the best possible learning rate for PAPN via grid-search. The results of this search are compiled in the second part of Table 2. Results in bold indicate highest performance and results underlined indicate second-best performance. It can be observed that by using a learning rate of 3e-5, better performance can be obtained, resulting in both a bigger gap in performance compared to other supervised methods as well as a reduction of the gap to close in on performance shown by DRL4Route without resorting to reinforcement learning for model training. This highlights that using a proper learning rate can be a good method to attain high performance.

## 5. Conclusion and Perspectives

To summarize the contributions in this work, a novel approach for route prediction is proposed by introducing proximity attention as a mechanism to model local relationships between jobs and locations through reachability mask exploitation. The proximity attention is then mixed with a global embedding of the problem to be fed to a decoder for route prediction node by node. The method is benchmarked against the other methods in literature and is the best performing supervised method in the case of information rich datasets. The model, although falling short of the reinforcement learning based method, is still competitive and poses the cost-return question when comparing both methods. As underlined earlier in this work, PAPN shows both higher epoch efficiency and time efficiency than DRL4Route (2023). As far as extensions go, this model will be integrated into a joint time-route prediction model and using the rich embeddings already available within the model for time prediction is promising. Researching cross-city transferability of supervised methods in that context is also interesting as all results reported earlier have training and testing sets from instances in the same city. Added to that is the idea of moving from a prediction model to an optimization model which remains a big challenge when working with supervised models as datasets of optimal data the size of LaDE (2024) are not yet readily available.